\pdfoutput=1
\documentclass[11pt]{article}

\usepackage[final]{coling}

\usepackage{times}
\usepackage{latexsym}
\usepackage{lettrine}
\usepackage{algorithm}
\usepackage{algorithmic}
\usepackage{amssymb}
\usepackage{siunitx}

\usepackage[T1]{fontenc}
\usepackage[utf8]{inputenc}

\usepackage{microtype}

\usepackage{inconsolata}

\usepackage{graphicx}

\usepackage{times}
\usepackage{url}
\usepackage{latexsym}
\usepackage{amsmath}
\usepackage{enumitem}
\usepackage{tabularx}
\usepackage{booktabs}
\usepackage{multirow}
\usepackage{float}
\usepackage{wrapfig}
\usepackage{placeins}
\usepackage{subcaption}
\usepackage{comment}
\usepackage{multicol}

\title{Comparative Analysis of Static and Contextual Embeddings for Analyzing Semantic Changes in Medieval Latin Charters}

\author{Yifan Liu\thanks{Corresponding author}, Gelila Tilahun, Xinxiang Gao, Qianfeng Wen, Michael Gervers\\
University of Toronto \\
\texttt{yifanliu.liu@mail.utoronto.ca}}

\begin{document}
\maketitle
\begin{abstract}
The Norman Conquest of 1066 C.E. brought profound transformations to England's administrative, societal, and linguistic practices. The DEEDS (Documents of Early England Data Set) database offers a unique opportunity to explore these changes by examining shifts in word meanings within a vast collection of Medieval Latin charters.  While computational linguistics typically relies on vector representations of words like static and contextual embeddings to analyze semantic changes, existing embeddings for scarce and historical Medieval Latin are limited and may not be well-suited for this task. This paper presents the first computational analysis of semantic change pre- and post-Norman Conquest and the first systematic comparison of static and contextual embeddings in a scarce historical data set. Our findings confirm that, consistent with existing studies, contextual embeddings outperform static word embeddings in capturing semantic change within a scarce historical corpus.
\end{abstract}

\section{Introduction}

The Norman Conquest of 1066 is a pivotal event in English history, marked by the introduction of new administrative and cultural practices by the Normans. This transformation is evident in the Medieval Latin charters --- official documents recording legal agreements, grants, rights, and privileges --- preserved in the DEEDS (Documents of Early England Data Set) corpus \cite{gervers2018dating}. One implication of these transformations is the shift in language usage and word meanings within the Medieval Latin charters, illustrated by the following examples: \textit{comes} generally meant “official” in Anglo-Saxon charters, but in Norman documents, it consistently appeared as a title meaning “earl” or “count”; \textit{proprius} (“one’s own”) was used by the Anglo-Saxons to indicate signing a document “with one’s own hand,” whereas the Normans used it to refer to property ownership. Investigating these changes in word meanings before and after the Norman Conquest --- a process known as lexical semantic change (LSC) --- provides insights into the cultural and societal transformations while also posing challenging research questions on how to systematically model this change.

In the field of computational linguistics, various methods have been proposed for modeling lexical semantics and thereby for studying semantic changes. In earlier years, static word embedding approaches, where each word was mapped to a fixed vector representation based on its co-occurrence patterns with other words within a corpus \cite{mikolov2013distributed, bojanowski2017enriching}, were dominant and proven effective in LSC studies \cite{kim2014temporal, hamilton2016diachronic}. In more recent years, contextual representations, which provide different vectors for the different contexts in which a word appears \cite{bert, elmo}, have achieved state-of-the-art performance in LSC studies, likely due to their ability to handle phenomena like polysemy and homonymy more effectively than static representations \cite{martinc2019leveraging, giulianelli2019lexical, kutuzov2022contextualized}.

Despite the successes of contextual embeddings in LSC research, they are typically trained on large corpora \cite{coha, googlebook} and require significantly more training data than static embeddings due to their more complex architectures and larger parameter sizes \cite{bommasani2021opportunities}. This poses a challenge for studies involving smaller data sets such as the DEEDS Medieval Latin corpus, which contains only 17k charters and 3M tokens --- considerably smaller than the billion-token corpora typically used to train contextual embeddings \cite{coha, googlebook}. Meanwhile, the Medieval Latin charters contain a rich and expansive vocabulary, including local dialects and borrowings from other languages (e.g., the Anglo-Saxon manuscripts include an extensive amount of Old English). These factors collectively raise concerns about the adaptability and relative performance of existing embedding methods in this scarce and heteroglossic data set.

Therefore, this paper aims to address the research gap in Medieval Latin charters with the following contributions: 
\begin{itemize}
    \item We present the \textbf{first LSC study} on \textbf{Medieval Latin charters} from England to understand the semantic change induced by the Norman conquest. These English Latin charters are exclusively a collection of legal documents pertaining to property rights whose topic and genre are quite different from other medieval Latin corpora described in section~\ref{othercorpora}.
    \item We provide a \textbf{systematic comparison} between static embeddings and contextual embeddings in modeling semantic change within Medieval Latin charters, which offers insights into the adaptability of these models within the context of a scarce and heteroglossic corpus.
\end{itemize}

The rest of this paper is organized as follows. Section \ref{sec:literature} summarizes the previous literature on static and contextual embeddings. Section \ref{sec:data} provides a detailed introduction to the DEEDS data set. Section \ref{sec:methods} outlines the training process for the different embedding methods on this corpus.\footnote{Corpus and codes available at: \url{https://anonymous.4open.science/r/historical-text-embedding-C328/README.md}} Sections \ref{sec:semantic} and \ref{sec:results} present the experiments, results, and discussions related to evaluating these embedding methods in capturing semantic change.

\section{Related Work}\label{sec:literature}
The standard computational approach for lexical semantic change (LSC) analysis involves separately training distributional representations, also known as embeddings, for different periods within a corpus \cite{gulordava2011distributional}, and then measuring the distance between the representations of a given word across these periods. In this section, we review the current approaches of semantic change analysis using static and contextual embeddings, explore their potential applications to Medieval Latin corpora, and identify gaps in the existing literature related to semantic change analysis in Medieval Latin.

\subsection{Static Word Embeddings}
Early approaches in the literature for learning word representations relied heavily on co-occurrence count-based methods \cite{lsa, count1}. With the advent of deep neural networks, prediction-based approaches gained popularity, including models such as the \textbf{Continuous Bag-of-Words} model \cite{mikolov2013distributed}, which encodes the contextual information of target words by predicting them based on the surrounding context; the \textbf{Continuous Skip-gram} model \cite{mikolov2013distributed}, which predicts surrounding context words based on the target word; and the \textbf{Subword model} \cite{bojanowski2017enriching}, which enhances these methods by learning context vectors through subword tokenization. 

The first research to integrate these prediction-based word vectors in LSC studies was conducted by \citet{kim2014temporal}. Subsequent work by \citet{hamilton2016diachronic} provided empirical evidence that neural-based diachronic embedding methods surpass traditional occurrence matrix-based approaches. Later studies have further refined these methods by incorporating subword models to enhance representation quality \cite{xu-etal-2019-treat,xu2021historical}.

In LSC, it is crucial to ensure that the embedding spaces for different periods are aligned to facilitate meaningful semantic change calculations. One effective strategy involves sharing the initial training weights of word embeddings, known as weight initialization, across different periods. \citet{kim2014temporal} introduced the \textbf{incremental initialization} approach, which initializes each year's training weights with vectors from the previous year.
\citet{montariol2019empirical} proposed two additional initialization strategies for scarce corpora: \textbf{internal initialization}, which trains a base model on the entire corpus and then incrementally trains embeddings for each period, and \textbf{backward external initialization}, which uses pre-trained embeddings from an external source to initialize the last time slice and then trains embeddings in reverse order. These approaches help to align embeddings across periods while also addressing data scarcity by leveraging large training corpora or external data sources, making them potentially well-suited for the scarce Medieval Latin charters.

\subsection{Contextual Embeddings}

Unlike static word embeddings that generate a single, fixed vector representation for each word, contextual embeddings produce unique representations for each instance of word usage context across a corpus, with \textbf{BERT} (Bidirectional Encoder Representations from Transformers) \cite{bert} being the most prominent example. Researchers such as \citet{hu2019diachronic}, \citet{giulianelli2019lexical}, and \citet{martinc2019leveraging} were among the first to incorporate these contextual word representations into lexical semantic change (LSC) studies. For instance, in \citet{martinc2019leveraging}, a pre-trained BERT model was further trained on another corpus (Liverpool FC) to understand semantic change, and the resulting word embeddings were aggregated to represent all instances of a word within a specific time-slice subcorpus. Since then, many studies have achieved strong results in LSC using contextual embeddings, applying these methods to various languages beyond English \cite{kanjirangat2020sst,rodina2021elmo,montariol2021measure,kurtyigit2021lexical,kutuzov2022contextualized}.

Most existing contextual representations are based on large, contemporary corpora, with limited attention given to low-resource historical corpora. Among the few works in this area, \citet{qiu2022histbert} explored the potential biases in using contemporary-trained contextual embeddings for historical data. They introduced histBERT, a model adapted from a base BERT model on the historical American English corpus (COHA), and demonstrated that histBERT outperforms the original BERT model in detecting semantic changes from historical texts. An alternative approach involves training BERT models from scratch, rather than adapting existing models, on historical data.
\citet{manjavacas-arevalo-fonteyn-2021-macberth} demonstrated the training-from-scratch approach with MacBERTh, a model trained on historical English corpora from 1450 to 1900 C.E., and showed that it outperformed the approach of adapting existing models in detecting semantic changes in historical English. Subsequent research extended the training-from-scratch approach to the German language tracing back to 750 C.E. by developing a GHisBERT model \cite{beck2023ghisbert}. These methods provide a foundation for developing contextual embeddings for Medieval Latin, a scarce and historical language.

\subsection{Towards Medieval Latin Embeddings}
\label{othercorpora}
Training word embeddings for Medieval Latin presents unique challenges due to a limited size of training corpora when compared to contemporary and modern languages. Several efforts have been made to construct Medieval Latin corpora to improve embedding training. Notable examples include the Dictionary of Medieval Latin from British Sources \cite{medievaldict}, which documents the Latin vocabulary used in Britain from 540 to 1600 C.E; Index Thomisticus, a digital corpus of Thomas Aquinas's 13th-century works \cite{mlatin1}; the Polish Medieval Latin Lexicon \cite{polish}, covering the 10th to mid-15th centuries; and the Frankfurt Latin Lexicon \cite{mehler2020frankfurt}, spanning the 6th to 9th centuries. These efforts have facilitated the development of high-quality static Latin word embeddings using CBOW, Skip-gram, and subword models. However, the topics and genres on which they focus differ from the DEEDS corpus in that DEEDS corpus is a collection of legal charters which primarily focuses on the rights of ownership and transfer of properties within Anglo-Saxon and Norman periods, which are critical sources for understanding impacts of the Norman conquest.

Contextual embeddings are believed to require even larger corpora, making their training on Medieval Latin languages more challenging than static embeddings. Although no contextual embeddings have been directly trained on Medieval Latin, some works have focused on Latin more broadly: \citet{bert} introduced Multilingual BERT, trained on the Wikipedia corpus for over 100 languages, including Latin; \citet{bamman2020latin} trained a BERT model specifically for Latin on a vast corpus of 600M tokens spanning from 200 B.C.E. to the present; Luis A. Vasquez trained a Latin BERT model on the Classical Language Toolkit (CLTK) corpus.\footnote{\url{https://huggingface.co/LuisAVasquez/simple-latin-bert-uncased}}

The historical language change of Latin has long attracted scholarly interest, and with the development of Latin corpora and word embeddings, researchers can now understand these changes computationally. For example, \citet{latinsemantic} analyzed Latin language change between the Classical and Medieval/Christian eras and evaluated different Latin embeddings on this task; \citet{romanlaw} detected semantic split in words with general and legal meanings by building Latin word embeddings from a 6th-century Roman law sourcebook; and SemEval 2020 \cite{semeval} included a task to calculate semantic change between the pre-Christian and Christian eras, using carefully annotated data from the LatinISE corpus \cite{latinise}.

However, significant research gaps still remain in the analysis of semantic change in Medieval Latin. First, there has been no computational evaluation of semantic change in the context of the Norman Conquest, a period marked by substantial administrative, cultural, and linguistic shifts \cite{gervers2018dating}. Second, although contextual embeddings have proven more powerful than static embeddings in large contemporary corpora, there is a lack of contextual embeddings specifically trained on scarce and historical Medieval Latin corpora, so a systematic comparison between these approaches is still needed.

% Although less extensive, there have been efforts to build MeLatin word embeddings, including Facebook's FastText \cite{fasttext}, trained on Wikipedia and Common Crawl, and the CoNLL shared tasks \cite{cnll}, trained on Common Crawl data. However, these data sources rely on automatic language detection engines to identify Latin text, which introduces significant noise due to the inclusion of other languages and contemporary terms. 

% Key contributions include the  corpus \cite{latinise}, spanning from 200 B.C.E. to the 20th century with 9 million tokens across diverse genres; Bamman et al.'s Latin book corpus \cite{bamman2012extracting}, sourced from the Internet Archive, with 1.4 billion tokens covering classical, medieval, and neo-Latin texts. 

\section{Data}\label{sec:data}
For our analysis, we used Medieval Latin charters from DEEDS (Documents of Early England Data Set).\footnote{\url{https://deeds.library.utoronto.ca/content/about-deeds}} The DEEDS database contains transcripts of over 70K Latin charters from the 7th to the 14th century. Of these, 40K pertain to England, and 17k are dated. They are official documents issued by kings and commoners and deal with the transfer of property and property rights. 

In this study, we focused on the 17k dated charters, as the dates were essential for splitting the corpus for semantic change analysis. We split the corpus into three sets:
the Anglo-Saxon period (from 589 to 1066 CE), referred to as \textbf{ANG} in later sections; the Norman period (from 1066 to 1153 CE), referred to as \textbf{NOR}; the later post-conquest period up to 1272 CE (also called Plantagenet period), referred to as \textbf{PLA}. Table~\ref{tab:corpus_stats} provides a summary of the corpus data.

\begin{table}[htbp]
    \centering
    \resizebox{0.48\textwidth}{!}{%
    \begin{tabular}{@{}lccc@{}}
        \toprule
        & \textbf{ANG} & \textbf{NOR} & \textbf{PLA} \\ 
        \midrule
        Time Span & 589-1065 & 1066-1153 & 1154-1272\\ 
        \# of Charters & 1432 & 4050 & 12926 \\
        \# of Tokens & 0.49M & 0.76M & 2.80M \\ 
        \bottomrule
    \end{tabular}
    } % Close resizebox
    \caption{Overview of the Medieval Latin corpus}
    \label{tab:corpus_stats}
\end{table}

The main focus of this paper is the semantic change induced by the Norman conquest (i.e., the transition from \textbf{ANG} to \textbf{NOR} periods, referred to as \textit{AN} in the later section). For comparison, we also examine the transitions from \textbf{NOR} to \textbf{PLA}, referred to as \textit{NP}.

\section{Models}\label{sec:methods}

\subsection{Static Word Embeddings}

We used the Continuous Skip-gram model with subword information \cite{mikolov2013distributed,bojanowski2017enriching}, as implemented in the FastText module in the Gensim library \cite{rehurek_lrec}, to generate static word embeddings for each period. We adopted the incremental initialization from \citet{kim2014temporal} as well as internal and backward external initialization from \citet{montariol2019empirical}. Due to resource constraints, we only tuned the embedding sizes (100 and 300) and the number of training epochs (10, 30, and 50) for each period and reported the best results.\footnote{See Appendix \ref{app:a} for details} All other hyperparameters were kept at their default settings in the FastText module.

\textbf{Incremental Initialization}: The embeddings from the previous period were used to initialize the embeddings for the subsequent period (incrementally). We refer to this model as \texttt{Incremental} in later sections.

\textbf{Internal Initialization}: We trained a base model on the full corpus for 50 epochs, which was then used to initialize the embeddings for the first period, with subsequent period embeddings being updated incrementally. We refer to this model as \texttt{Internal} in later sections.

\textbf{Backward External Initialization}: We utilized pre-trained Latin word embeddings from \citet{grave2019unsupervised} on Common Crawl and Wikipedia as the base model. Then, we incrementally updated each period's embeddings from the most recent to the oldest, a reverse updating process that might be beneficial to our corpora, which have lower volumes in the older periods \cite{montariol2019empirical}.\cite{montariol2019empirical}. We refer to this model as \texttt{External} in later sections.

\subsection{Contexual Embeddings}
\textbf{BERT Trained from Scratch}: 
We pre-trained a BERT model from scratch on the full Medieval Latin charters corpus using the hyperparameters recommended by \citet{manjavacas2022adapting} in historical English and \citet{beck2023ghisbert} in historical German. The model consists of 12 hidden layers, each with 768-dimensional embeddings, and 12 attention heads, with a vocabulary size of 32,000 tokens. Training was conducted over 10 epochs with a batch size of 8 using the masked language modeling (MLM) task, where 10\% of the tokens were randomly masked. We refer to this model as \texttt{MLatin-BERT} in later sections.
 
\textbf{BERT Adapted from Pre-trained Models}: For comparison, we continued training two Latin BERT models on the Medieval Latin charters corpus: the first, Latin-BERT by \citet{bamman2020latin} \footnote{\url{https://github.com/dbamman/latin-bert}}, which was trained on a diverse range of Latin corpora with 600M tokens spanning from 200 B.C.E. to the present, and the second, \texttt{simple-latin-bert-uncased} by Luis A. Vasquez \footnote{\url{https://huggingface.co/LuisAVasquez/simple-latin-bert-uncased}}, which was trained using corpora from the Classical Language Toolkit (CLTK). Both models were configured with standard BERT hyperparameters with a hidden size of 768 and 12 layers. They were further trained from their last checkpoints on the Medieval Latin corpus for an additional 4 epochs, as recommended by the original BERT paper \cite{bert}. We refer to these models as \texttt{Ada-BERT-Bam} and \texttt{Ada-BERT-Vas}, respectively, in later sections.

\textbf{Tokenizer}: We pre-trained a tokenizer for all described models, which accounts for the diverse word forms in the Medieval Latin charters. The tokenizer was trained with the same hyperparameter settings outlined by \citet{beck2023ghisbert} using the HuggingFace \texttt{BertWordPieceTokenizer} module with a vocabulary of 32000 and a maximum sequence length of 512. 

\textbf{Extract Word Embeddings}:
To enable direct comparison between contextual and static embeddings in the semantic change analysis, we followed the method described by \citet{martinc2019leveraging} to extract word embeddings from contextual embeddings for each time period (discussed in Section \ref{sec:data}), as detailed in Algorithm \ref{alg:lsc_embeddings}.

\begin{algorithm}[t]
\caption{Extract and average word embeddings from contextual embeddings for a time period}
\label{alg:lsc_embeddings}
\textbf{Input}: Medieval Latin texts for a given time period, \(\mathcal{C} = \{S_1, S_2, \ldots, S_n\}\), where \(S_i\) is a sentence. Contexual embeddings \(\mathcal{E} = \{\mathbf{E}_{S_1}, \mathbf{E}_{S_2}, \ldots, \mathbf{E}_{S_n}\}\), where \(\mathbf{E}_{S_i} \in \mathbb{R}^{L \times d}\) is the embedding matrix for sentence \(S_i\). \\
\textbf{Output}: Word embeddings \(\mathbf{W} \in \mathbb{R}^{M \times d}\), where \(M\) is the number of distinct words in \(\mathcal{C}\).

\begin{algorithmic}[1]
\STATE Initialize word embedding matrix \(\mathbf{W}\)
\FOR{each distinct word $w_j \in \mathcal{C}$}
    \STATE Initialize embedding sets \(\mathcal{W}_j = \{\}\)
\ENDFOR
\FOR{each sentence \( S_i \in \mathcal{C} \)}
    \STATE \(\mathbf{S}_i \leftarrow \frac{1}{4} \sum_{l=L-3}^{L} \mathbf{E}_i^{(l)}\) \COMMENT{Compute sentence embedding using last four layers}
    \FOR{each word \( w_j \in S_i \)}
        \STATE Identify the word pieces \(\mathbf{P}_j\) corresponding to word \( w_j \) using offset mappings.
        \STATE Compute word embedding: \(\mathbf{w}_j^{(S_i)} \leftarrow \frac{1}{|\mathbf{P}_j|} \sum_{p \in \mathbf{P}_j} \mathbf{S}_i^{(p)}\) \COMMENT{Compute word embedding for \( w_j \) in sentence context \( S_i \)}
        \STATE Store \(\mathbf{w}_j^{(S_i)}\) in set \(\mathcal{W}_j\)
    \ENDFOR
\ENDFOR
\FOR{each word \( w_j \) in vocabulary}
    \STATE \(\bar{\mathbf{w}}_j \leftarrow \frac{1}{|\mathcal{W}_j|} \sum_{\mathbf{w}_j^{(S_i)} \in \mathcal{W}_j} \mathbf{w}_j^{(S_i)}\) \COMMENT{Compute average embedding}
    \STATE Store \(\bar{\mathbf{w}}_j\) in \(\mathbf{W}\)
\ENDFOR
\STATE return $\mathbf{W}$
\end{algorithmic}
\end{algorithm}

\section{Lexical Similarity Analysis}\label{sec:semantic}

\subsection{Similarity Measures}

To evaluate the applicability of different embedding models in analyzing semantic change within the Medieval Latin charters, we conducted a semantic similarity analysis across various periods following the approach of \citet{beck2023ghisbert}. Specifically, for a given word \( w \) occurring in two periods \( t_1 \) and \( t_2 \), we computed the cosine similarity between their embeddings \(\mathbf{w}_{t_1}\) and \(\mathbf{w}_{t_2}\) using the following formula:

\begin{equation}
\text{Cos}(\mathbf{w}_{t_1}, \mathbf{w}_{t_2}) = \frac{\mathbf{w}_{t_1} \cdot \mathbf{w}_{t_2}}{\|\mathbf{w}_{t_1}\| \|\mathbf{w}_{t_2}\|}
\end{equation}

A lower cosine similarity score between periods suggests a potential semantic shift in the word’s meaning \cite{kim2014temporal, giulianelli2019lexical}.

In our analysis, we divided the data into three periods, as outlined in Section \ref{sec:data}, and therefore, for each word, we computed two cosine similarity measures: \(\text{COS}_{AN}\), representing the transition from \textbf{ANG} to \textbf{NOR} and \(\text{COS}_{NP}\), representing the transition from \textbf{NOR} to \textbf{PLA}. We will refer to the above labels in later sections. 

\subsection{Data Set Labeling} \label{sec: label}
To quantitatively assess the performance of different embedding methods, we applied the following labeling procedure to the data set.  We selected commonly occurring words with a relative frequency exceeding five occurrences per 100,000 in all periods, resulting in 662 words in total. Three Latin specialists with domain knowledge were asked to make a binary decision on whether the meaning of each word had changed from the Anglo-Saxon to the Norman period (marked as 1) or remained unchanged (marked as 0), which were then used as \textbf{semantic change labels} for subsequent studies. For each period, the labelers made their decisions on a word by reviewing 10 sample sentences containing the word. If all three labelers agreed on a label, the word was classified as either \textit{changed} (for positive cases, 41 words) or \textit{unchanged} (for negative cases, 297 words)\footnote{The list of changed and unchanged words can be found at: \url{https://anonymous.4open.science/r/historical-text-embedding-C328/README.md}}. Examples of \textit{changed} words include \textit{finis}, which shifted from meaning "end" or "completion" in Anglo-Saxon times to "fine" as a payment in a final agreement in Norman, and \textit{honorifice}, which originally meant "honorable" or "honorably" in the context of a king's duties, but in Norman documents referred specifically to the manner in which land was held by a feudal lord. 
Examples of \textit{unchanged} words include pronouns (e.g., \textit{meus}, "my"), numbers (e.g., \textit{centum}, "hundred"), greetings (e.g., \textit{salute}, "hello"), and prepositions (e.g., \textit{post}, "after"; \textit{usque}, "until"). In cases where no consensus was reached, the words were excluded from both categories. Our analysis focused solely on the 338 target words that were clearly categorized as either \textit{changed} or \textit{unchanged}.
\begin{table*}[htbp]
\centering
\begin{tabularx}{\textwidth}{@{}cc *{6}{S[table-format=-1.3,table-space-text-post=*]}@{}}
\toprule
& & \multicolumn{3}{c}{\textbf{Static}} & \multicolumn{3}{c}{\textbf{Contextual}} \\ \cmidrule(lr){3-5} \cmidrule(lr){6-8}
& & \texttt{Incremental} & \texttt{Internal} & \texttt{External} & \texttt{MLatin-BERT} & \texttt{Ada-BERT-Bam} & \texttt{Ada-BERT-Vas} \\
\midrule
\multirow{2}{*}{\textit{\textbf{AN}}} & $\delta_{\mu}$ & 0.054* & -0.004 & 0.002 & 0.047* & 0.037* & 0.055* \\
                    & $\rho$ & -0.169* & 0.018 & -0.120 & -0.481* & -0.395* & -0.360* \\
\midrule
\multirow{2}{*}{\textit{\textbf{NP}}} & $\delta_{\mu}$ & 0.011 & -0.015 & -0.003 & 0.009 & 0.006 & 0.012 \\
                    & $\rho$ & -0.003 & 0.055 & -0.072 & -0.135* & -0.126* & -0.141* \\
\bottomrule
\end{tabularx}
\caption{Quantitative results of static and contextual embeddings in semantic for the \textit{AN} and \textit{NP} periods. Two metrics are reported: $\delta_{\mu}$ indicates the difference in mean cosine similarity between the \textit{unchanged} and \textit{changed} word groups, and $\rho$ represents the correlation between semantic change labels and cosine similarity measures for each target word across two periods. An asterisk (*) denotes statistically significant results ($t$-test, $p < 0.01$).}
\label{tab:sec6.1}
\end{table*}

\begin{figure*}[h!]
    \centering
    \begin{subfigure}[b]{1\linewidth}
    \includegraphics[width=\linewidth, height=0.5\linewidth]{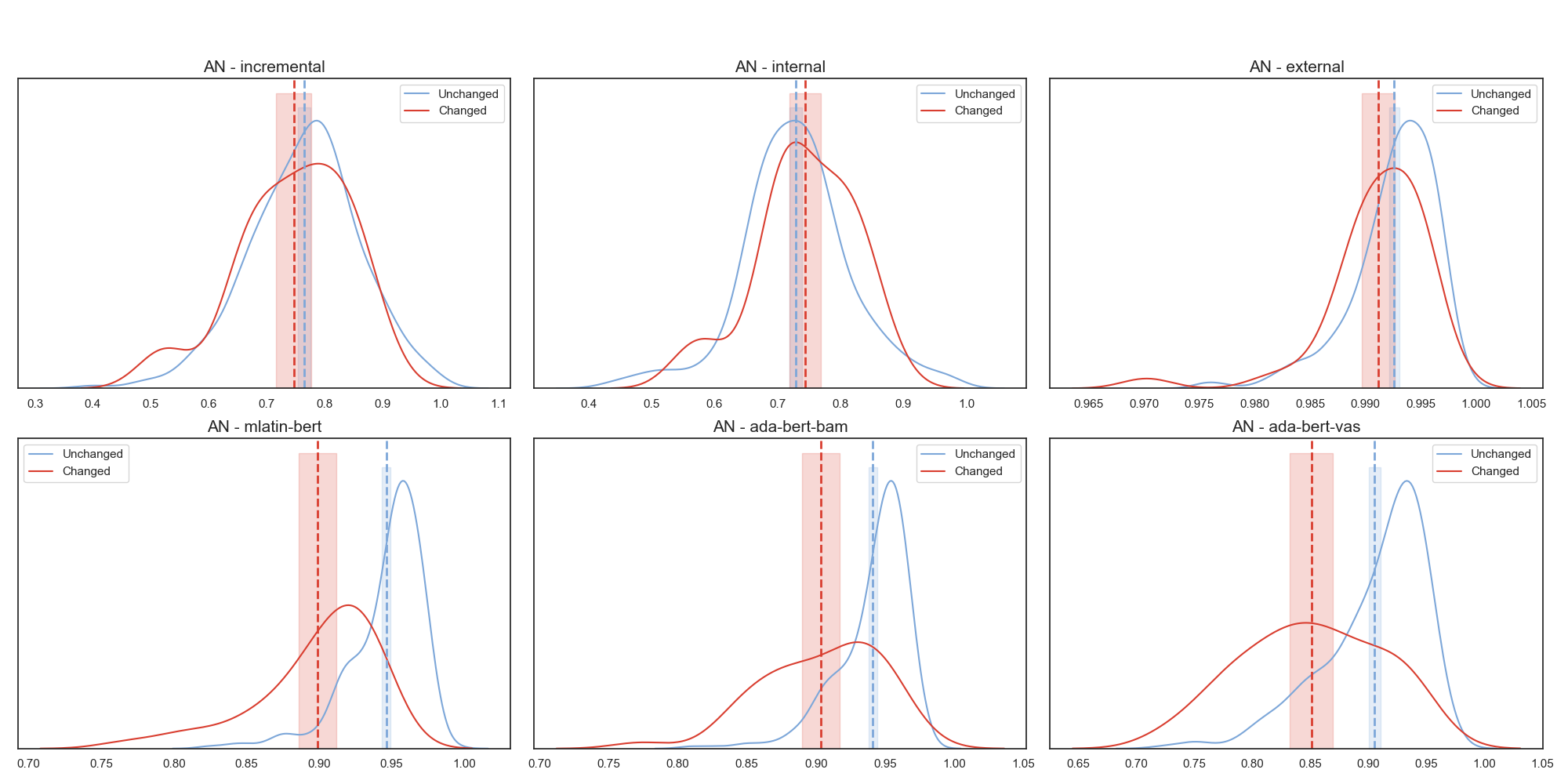}
        \caption{\textit{AN} period }
        \label{fig:AN-intra}
    \end{subfigure}
    
    \begin{subfigure}[b]{1\linewidth}
    \includegraphics[width=\linewidth, height=0.5\linewidth]{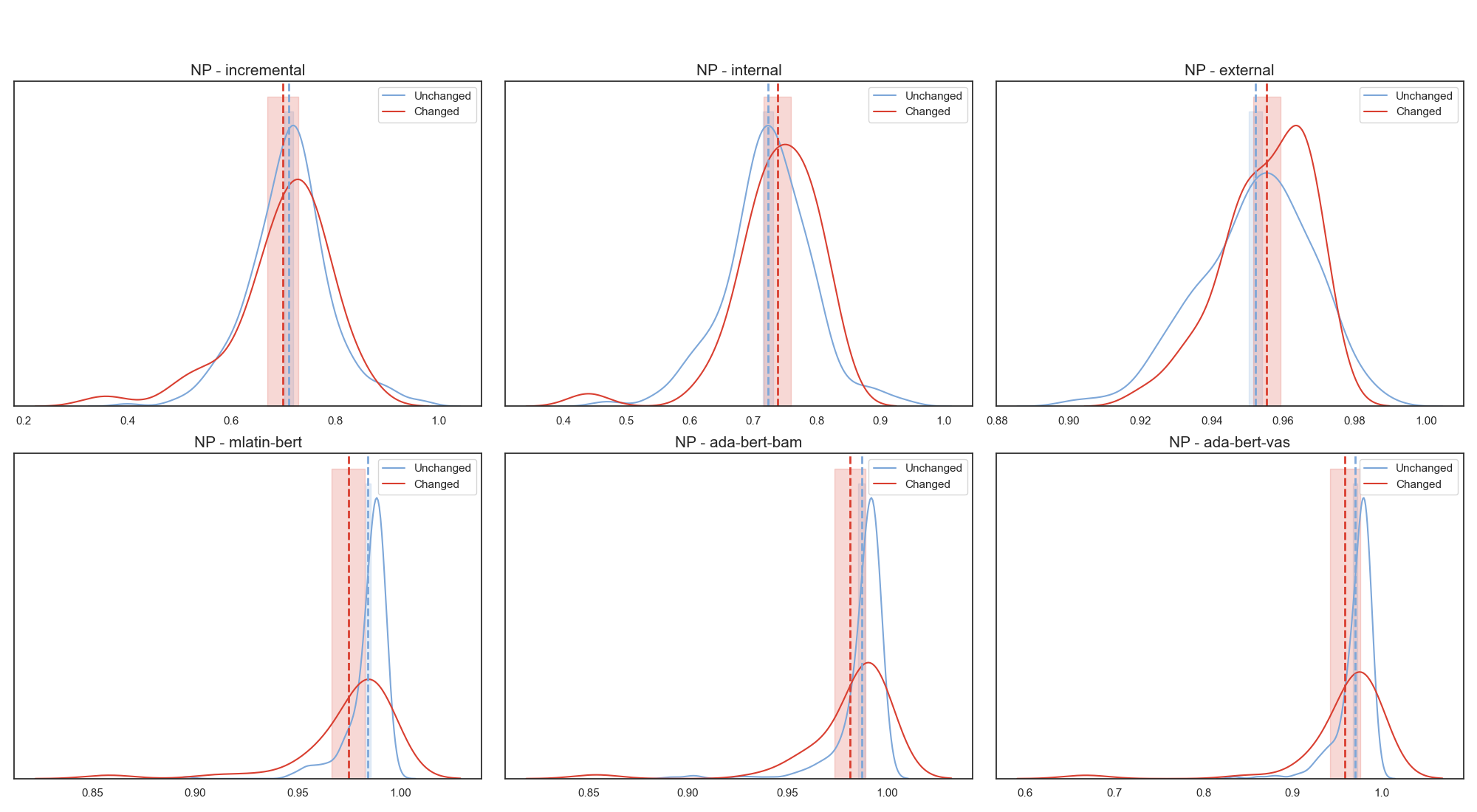}
        \caption{\textit{NP} period }     
        \label{fig:NN-intra}
    \end{subfigure}
    \caption{Distribution of cosine similarity for \textit{changed} and \textit{unchanged} words across different embedding models -- \textit{AN} period (top) and \textit{NP} period (bottom). The dashed lines represent the mean cosine similarity for \textit{changed} and \textit{unchanged} words across the two periods and for each model. The shaded areas represent the 95\% confidence intervals.}
\end{figure*}

\section{Results} \label{sec:results}

\subsection{Semantic Change in \textit{AN} Period} \label{sec: res AN}

Given our primary focus on the semantic changes induced by the Norman Conquest, we first present the results of $\text{COS}_{AN}$ (i.e., the cosine similarity between the embeddings from the Anglo-Saxon and Norman periods for a given word) across different embedding models (as discussed in Section \ref{sec:methods}). The \textit{AN} section of Table \ref{tab:sec6.1} reported two performance metrics: the difference in the averages of the $\text{COS}_{AN}$ between \textit{unchanged} and \textit{changed} words (as discussed in Section \ref{sec:methods}), $\delta{\mu}$, where a larger difference indicates a better ability to distinguish between the two groups; the Pearson correlation, $\rho$, between the binary change labels and $\text{COS}_{AN}$ for all target words, with values ranging from -1 (strong negative correlation, the most desirable outcome) to 1 (strong positive correlation, the least desirable outcome). All contextual embeddings demonstrated statistically significant $\delta_{\mu}$ values. The correlation coefficient further highlighted the better performance of contextual embeddings in semantic change analysis, with \texttt{MLatin-BERT} achieving the strongest negative correlation ($\rho = -0.481$) and outperforming models adapted from pre-trained Latin BERT. Among the static embedding methods, \texttt{Incremental} and \texttt{External} showed fair results, with the correct direction of $\delta_{\mu}$ and a moderate negative correlation between true semantic change labels and cosine similarity, although the correlation was much weaker than that of the contextual models. In contrast, \texttt{Internal} produced results opposite to those expected.

Figure \ref{fig:AN-intra} displays a more detailed distributions, mean values, and 95\% confidence intervals of $\text{COS}_{AN}$  for both the \textit{changed} and \textit{unchanged} word groups. Contextual embeddings consistently showed an obvious difference between the distributions of \textit{changed} and \textit{unchanged} words, with \textit{changed} words centering around much lower cosine similarity scores.  Notably, \texttt{Ada-BERT-Vas} produced lower similarity for both word groups compared to \texttt{MLatin-BERT} and \texttt{Ada-BERT-Bam}. The results for static embeddings reveal several concerns: while \texttt{Incremental} identified the correct difference in mean values (with the mean cosine similarity being smaller for the \textit{changed} word group), it did not show a significant difference in the distribution shapes between the two word groups. The \texttt{External} model exhibited a difference in distribution, but the absolute difference in mean cosine similarity was marginal (only around 0.002). The \texttt{Internal} approach produced completely opposite to the expected results.

Overall, these results suggest that contextual embeddings are more effective at capturing semantic changes and distinguishing \textit{changed} words from \textit{unchanged} words, even in a scarce and historical language setting, which demonstrates the adaptability of contextual embeddings to smaller data sets beyond what has been shown in existing literature. Additionally, we found that both static and contextual models trained from scratch (\texttt{Incremental} and \texttt{MLatin-BERT}) performed better than those adapted from pre-trained embeddings, likely due to the lack of high-quality base representations for Medieval Latin texts.

\subsection{Comparison Across Periods}
For comparison, we also report the distributions, $\delta_{\mu}$ between the \textit{unchanged} and \textit{changed} groups of $\text{COS}_{NP}$ (i.e., the cosine similarity between the embeddings from the Norman and Plantagenet periods for a given word), and the correlation $\rho$ between semantic change labels and $\text{COS}_{NP}$. We expect the \textit{AN} period to have a smaller mean value across all words, a larger mean difference between \textit{changed} and \textit{unchanged} words, and a more negative correlation between $\text{COS}_{NP}$ and semantic change labels than for \textit{NP} period, based on the assumption that the semantic change from the Anglo-Saxon period to the Norman period is more significant than 
from Norman to Plantagenet (often seen as a continuation of Norman ruling) due to the profound linguistic, cultural, and sociological shifts triggered by the Norman Conquest \cite{clanchy2012memory}. 

The results from Figure \ref{fig:NN-intra} indicate that all contextual embeddings find higher distribution center values for both \textit{changed} and \textit{unchanged} words during the \textit{NP} period than \textit{AN} period. Additionally, the \textit{NP} section of Table \ref{tab:sec6.1} reveals that LL contextual embeddings identify significantly larger $\delta_{\mu}$ and more negative $\rho$ during the \textit{AN} periods. These results suggest that contextual embeddings effectively differentiate periods of dramatic semantic change from relatively stable periods. Among the static embeddings, although the \texttt{Incremental} and \texttt{External} approaches correctly demonstrate smaller $\delta_{\mu}$ and weaker $\rho$ in the \textit{NP} period compared to the \textit{AN} period, they fail to capture the difference in absolute mean cosine similarity, as both models display lower mean cosine similarity across all word groups in the \textit{NP} period than in the \textit{AN} period.

\section{Conclusion}
This paper represents the first effort to explore semantic changes in the Medieval Latin charters as a result of the Norman Conquest, and the first to systematically implement and compare static and contextual word embeddings in the context of the scarce and historical corpus. Our evaluation on the DEEDS Medieval Latin charters corpus with manually labeled semantic changes demonstrates that contextual embeddings outperform static word embeddings, even on a scarce and complex historical data set. This finding is consistent with results from large contemporary data sets and confirms the adaptability of contextual embeddings to smaller data sets beyond what has been shown in existing literature. Furthermore, consistent with previous work on building contextual embeddings for historical corpora \cite{manjavacas-arevalo-fonteyn-2021-macberth, beck2023ghisbert}, training from scratch yields better performance in capturing the correlation between semantic change labels and similarity measures.

\section*{Future Work}
This research opens new avenues for historical linguistics by providing a framework to explore semantic change in Medieval Latin charters and understand the social, cultural, and political impacts of the Norman Conquest. One could utilize the semantic change analysis framework discussed in this paper as a knowledge discovery process to learn previously unrealized shifts in word meaning.

However, this study also faces certain limitations. As an initial exploration of diachronic embeddings in Medieval Latin charters, we lack a gold standard data set for semantic change detection and were only able to construct binary semantic change labels due to resource constraints. Future work could involve collaboration with more Medieval Latin scholars to develop a continuous semantic change index ranging from zero to one, which could allow for more informative and rigorous quantitative evaluations of our models and establish a benchmark for subsequent research in this field. Additionally, this study has primarily used cosine similarity between word embeddings from different periods as the metric for modeling semantic change, which may not be the most appropriate measure. Future research could explore alternative distance-based metrics, such as Average Pairwise Distance (APD) and Inverted Cosine Similarity over Prototypes (PRT), as suggested in previous studies \cite{giulianelli2020analysing,kutuzov2022contextualized}.

\begin{comment}
\section*{Acknowledgement}
Each member was responsible for different aspects of the research. Yifan Liu developed the overall idea of the project, implemented the code for the BERT model, and wrote the introduction, related work, and BERT-related methods sections. Feixuan Chen handled the intrinsic evaluation of the models and conducted the lexical semantic change detection, including both coding and writing the relevant sections of the paper. Xinxiang Gao focused on the static word embedding models, conducted a literature review on these methods, implemented the training procedures, and wrote the corresponding sections of the paper. Gangquan Zhang was responsible for the extrinsic evaluation of the models, wrote the relevant sections, and developed the code to extract static BERT models. We also thank Prof. Michael Gervers and Dr. Gelila Tilahun for providing us with access to DEEDS project database, sharing their knowledge of Medieval Latin from the Anglo-Saxon and Norman periods, and manually evaluating the inter-concept similarity test results.
\end{comment}

\bibliography{custom}

\appendix
\section{Hyperparameter Experiments for Static Embeddings} \label{app:a}
\begin{figure}[h!]
    \centering
    \begin{subfigure}[b]{1\linewidth}
        \includegraphics[width=\linewidth]{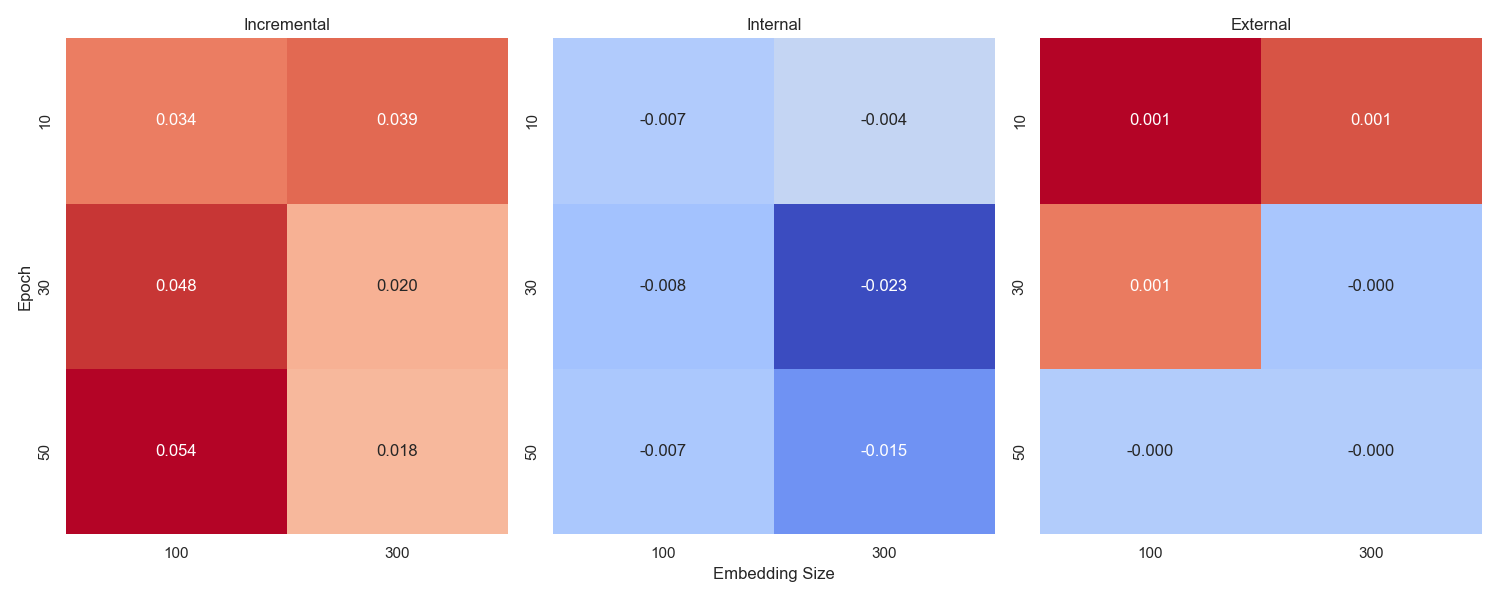}
    \end{subfigure}
    
    \begin{subfigure}[b]{1\linewidth}
        \includegraphics[width=\linewidth]{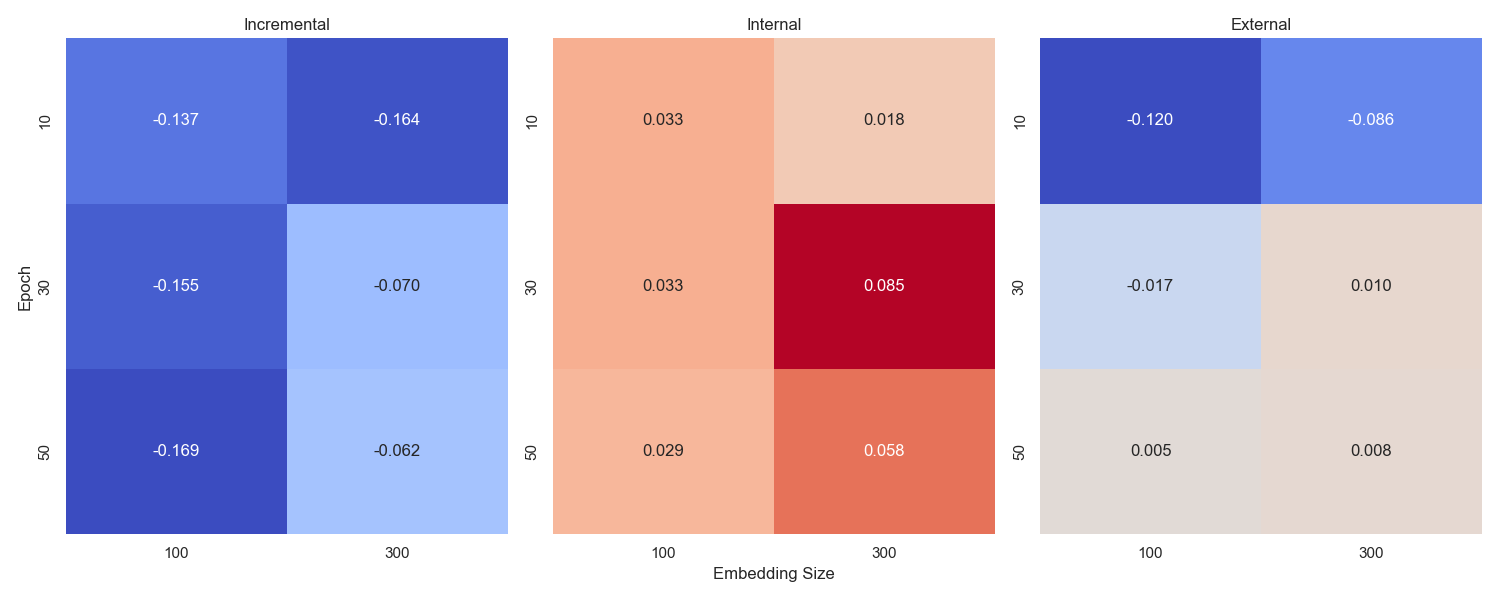}
    \end{subfigure}
    \caption{Heatmaps showing the evaluation metrics varying across different hyperparameter settings, with $\delta_{\mu}$ (top) and $\rho$ (bottom).}

    \label{fig: hyper}
\end{figure}

This section details the hyperparameter selection for static embeddings. Figure \ref{fig: hyper} illustrates how the evaluation metrics in \textit{AN} period, $\delta_{\mu}$ and $\rho$ (see detailed definitions and significance in Section \ref{sec: res AN}), vary across different hyperparameter settings, specifically the number of training epochs (10, 30, and 50) and the embedding size (100 and 300).

For the \texttt{Incremental} approach, the best hyperparameters were found when the embedding size was set to 100 and the number of training epochs was 50. A clear trend emerges where an embedding size of 100 outperforms a size of 300. Additionally, with a embedding size of 100, increasing the number of training epochs leads to better results, whereas with a embedding size of 300, fewer training epochs yield better outcomes.

In the \texttt{External} approach, the optimal hyperparameters were identified when the embedding size was 100 and the training epochs were set to 10. There is a trend indicating that smaller embedding sizes and fewer training epochs produce better results for this approach.

For the \texttt{Internal} approach, the best performance was observed when the embedding size was 300 and the number of training epochs was 10. However, the results do not exhibit a consistent trend across different hyperparameter settings and embedding sizes.

\section{Effect of Model Size on Contextual Embeddings}
In this section, we examine how the size of a BERT model trained from scratch affects performance during the \textit{AN} period. In addition to the \texttt{MLatin-BERT} model, we trained two smaller models: a small BERT model (4 attention heads, 4 hidden layers, and an embedding size of 256) and a medium BERT model (8 attention heads, 8 hidden layers, and an embedding size of 512), both of which are smaller than \texttt{MLatin-BERT}.\footnote{Future work could explore larger BERT models, which we did not pursue due to resource constraints.} As shown in Table \ref{tab:model size}, there is a clear trend where larger model sizes result in better performance, evidenced by the greater differences in mean cosine similarity and stronger correlations between the semantic change labels and cosine similarity for larger models. These findings are consistent with established scaling laws \cite{scaling}. %\cite{kaplan2020scaling}.

\begin{table}[h!]
\centering
\begin{tabular}{@{}l *{3}{S[table-format=-1.3]}@{}}
\toprule
& \texttt{Small} & \texttt{Medium} & \texttt{MLatin-BERT} \\
\midrule
$\delta_{\mu}$ & 0.012 & 0.028 & 0.047 \\
$\rho$         & -0.250 & -0.327 & -0.481 \\
\bottomrule
\end{tabular}
\caption{Evaluation metrics ($\delta_{\mu}$ and $\rho$) across different model sizes: Small, Medium, and Large (\texttt{MLatin-BERT}).} 
\label{tab:model size}
\end{table}

\section{Effect of Adaption on Contextual Embeddings}
In this section, we examine how adapting a pre-trained BERT model to Medieval Latin charters affects performance. We replicate the study for the \textit{AN} period using Latin-BERT \cite{bamman2020latin}. Table \ref{tab:ada} shows that domain adaptation of the pre-trained Latin BERT model to Medieval Latin charters enhances its ability to identify semantic change, as evidenced by the greater difference in mean cosine similarity and the stronger correlation between the semantic change labels and cosine similarity observed in the \texttt{Ada-BERT-Bam} model.

\begin{table}[h!]
\centering
\begin{tabular}{@{}l *{2}{S[table-format=-1.3]}@{}}
\toprule
& \texttt{Latin-BERT-Bam} & \texttt{Ada-BERT-Bam} \\
\midrule
$\delta_{\mu}$ & 0.020 & 0.037 \\
$\rho$         & -0.326 & -0.395 \\
\bottomrule
\end{tabular}
\caption{Evaluation metrics ($\delta_{\mu}$ and $\rho$) for \citet{bamman2020latin}'s Latin BERT (\texttt{Latin-BERT-Bam}) and the adapted version (\texttt{Ada-BERT-Bam}).} 
\label{tab:ada}
\end{table}

\end{document}